# The Vibe-Automation of Automation: A Proactive Education Framework for Computer Science in the Age of Generative AI

**Ilya Levin**

*Holon Institute of Technology, Holon, Israel*, levini@hit.ac.il

## Abstract

The emergence of generative artificial intelligence (GenAI) represents not an incremental technological advance but a fundamental epistemological rupture — a qualitative shift in the nature of computation that challenges the foundational assumptions of computer science as an academic discipline. Pedro Domingos famously defined machine learning as "the automation of automation"—a paradigm in which computers program the logic required to solve problems. This paper proposes the concept of "Vibe-Automation" to characterize the decisive shift that generative AI introduces: from optimizing objective metrics (accuracy, efficiency) to navigating subjective qualities (tone, style, intent, and contextual judgment).

The central philosophical claim is that the significance of GenAI does not reside primarily in the introduction of natural language interfaces or new forms of abstraction, but in its functional access to what we term operationalized tacit regularities — contextual, practice-embedded patterns traditionally associated with tacit knowledge. While generative systems do not possess tacit knowledge in a phenomenological sense, they operationalize sensitivities to context, nuance, and situated judgment that cannot be exhaustively specified through explicit algorithmic rules. We argue that "the vibe" is not merely a colloquialism but a functional description of high-dimensional vectors in latent space that encode semantic and emotional context. Consequently, the human role evolves from the "algorithmic thinker" to the "Vibe-Engineer" — a professional who manages resonance, alignment, and contextual coherence across multiple computational regimes.

On this basis, the article develops a two-dimensional conceptual--operational matrix that integrates epistemological analysis with educational transformation. Three levels of depth — conceptual foundations, transformation principles, and institutional implementation — are articulated across three interdependent pillars of action: faculty worldview, professional and industry relations, and curriculum structure. Finally, we address the risks of "mode collapse" and cultural homogenization (the "Beige Box" problem), arguing that mastering Vibe-Automation is essential to prevent a regression into averaged, synthetic mediocrity.

# 1. Introduction

Computer science stands at a critical inflection point. The rapid emergence of generative artificial intelligence — large language models, transformer-based architectures, and systems capable of producing code, text, and structured reasoning at unprecedented levels — has triggered changes that cannot be adequately addressed through incremental curricular adaptation. These changes are not confined to tools or techniques; they affect the epistemic foundations of computation itself and, consequently, the educational structures built upon them. This article develops a comprehensive framework for rethinking computer science education in response to what we argue is not merely a technological advance, but an epistemological rupture.

Historically, higher education has responded to technological change in an additive manner: new courses are introduced, existing syllabi are updated, and faculty acquire supplementary technical skills. Such responses presuppose continuity in the underlying nature of professional practice and disciplinary knowledge. The emergence of GenAI disrupts this presupposition. It introduces forms of computational behavior that are not fully explicable, controllable, or verifiable using the conceptual frameworks developed for symbolic, rule-based systems.

Pedro Domingos famously defined machine learning as "the automation of automation"—a formulation that captured the moment when computers began not merely executing human-written programs but generating them themselves (Domingos, 2015). Yet for all its elegance, this definition embedded an assumption that now requires scrutiny: the automation it described remained within the domain of explicit, rule-governed procedures. The machine learned to classify, predict, and decide — but it did so by discovering patterns expressible as mathematical functions operating on structured inputs. The automation was of the "how" —how to classify, how to predict, how to decide — while the "what" and the "why" remained firmly human prerogatives.

Generative AI does not simply automate the production of explicit rules from data. It produces outputs — texts, images, code, musical compositions, strategic analyses — that exhibit qualities traditionally attributed to human judgment, taste, and intuitive understanding. What is being automated is no longer merely the "how" of problem-solving but something closer to what Michael Polanyi termed "tacit knowledge": the dimension of competence that resists exhaustive formalization (Polanyi, 1966). We are witnessing a move from the automation of automation to what may be called the vibe-automation of automation.

The term "vibe," while colloquial, is used here as a precise technical descriptor. In the context of generative systems, "vibe" refers to high-dimensional contextual representations—vectors in latent space — that encode semantic, stylistic, and emotional regularities. These representations resist reduction to explicit features, yet they demonstrably guide system behavior in ways that parallel human tacit competence. Karpathy's (2025) concept of "vibe-coding" captures this shift at the practitioner level, but the transformation it signals is far more fundamental than a change in programming methodology.

A common response to generative AI situates it within a familiar historical narrative of increasing abstraction in programming: from machine code to assembly, from assembly to high-level languages, and from these to natural-language interaction. Section 2 argues that this continuity narrative is fundamentally misleading: previous increases in abstraction preserved the epistemic structure of explicit specification followed by deterministic execution, while generative AI introduces a qualitatively different regime.

Interaction with generative AI fundamentally alters this epistemic structure. In vibe-coding practices, users do not specify procedures to be executed; instead, they guide the behavior of systems whose internal processes are neither fully transparent nor deterministically controllable. The relevant form of expertise is not limited to the precision of instruction but extends to contextual judgment: the ability to frame tasks appropriately, evaluate outputs critically, recognize failure modes, and iteratively shape system behavior in the absence of full epistemic control (Sarkar et al., 2022). Vibe-coding should not be understood as the next stage of abstraction. It constitutes a distinct epistemic practice characteristic of a different computational regime — one in which collaboration, interpretation, and evaluation replace command and specification as central modes of engagement.

Computer science education thus faces a dual challenge: to preserve the theoretical rigor of its algorithmic foundations while simultaneously preparing students to operate across newly emergent computational regimes. To address this challenge, the article proposes the paradigm of proactive education as an organizing principle for curriculum transformation. The term is introduced here to distinguish the approach advocated from both reactive adaptation (responding to change after the fact) and preventive measures (attempting to insulate education from disruption). Proactive education shifts the focus of professional preparation from the acquisition of stable skills to the cultivation of sustained agency under conditions of structural uncertainty.

The argument proceeds as follows. Section 2 establishes the epistemological foundations by characterizing generative AI as an epistemological rupture and developing the concept of operationalized tacit regularities. Section 3 examines the technical and cognitive dimensions of vibe-automation, introducing the figure of the Vibe-Engineer and the framework of plural computational regimes. Section 4 presents the two-dimensional transformation matrix that integrates epistemic depth with institutional action. Section 5 elaborates on the transformation across three institutional pillars — faculty worldview, industry relations, and curriculum — with concrete illustrative examples. Section 6 addresses the risks of cultural homogenization inherent in generative optimization. Section 7 offers concluding reflections.

## 2. Generative AI as an Epistemological Rupture

The previous section argued that generative AI introduces a qualitative shift in the nature of computation — one that cannot be accommodated within existing educational frameworks. The present section develops this claim in three steps. First, it characterizes the nature of the epistemological rupture by contrasting the symbolic paradigm with the emergent behavior of generative systems (2.1). Second, it introduces the concept of operationalized tacit regularities as a philosophically precise way of understanding what generative AI does without resorting to anthropomorphic claims (2.2). Third, it illustrates the distinction between symbolic and vibe-based automation through a thought experiment about extraterrestrial communication, clarifying the operational logic of contextual coherence (2.3).

### 2.1. The Problem of Understanding: From Symbolic AI to Emergent Behavior

The most fundamental obstacle to effective engagement with generative artificial intelligence is not technical but conceptual. It consists in a widespread failure to recognize the nature of the transformation at hand. This failure reflects a deeper epistemological inertia rooted in long-standing assumptions about computation, artificial intelligence, and the forms of knowledge they can legitimately accommodate.

For the greater part of its history, artificial intelligence developed within the symbolic paradigm, or Good Old-Fashioned AI (GOFAI). Within this paradigm, intelligent behavior is understood as the manipulation of symbols according to explicitly defined rules (Haugeland, 1985). However complex such systems may become, their behavior remains, in principle, transparent, decomposable, and traceable to identifiable components and procedures. This paradigm shaped not only AI research but also the self-understanding of computer science as a discipline fundamentally oriented toward explicit knowledge.

Generative AI does not represent a refinement or culmination of this paradigm. It constitutes a departure from it. Contemporary generative systems — most notably large language models based on transformer architectures (Vaswani et al., 2017) — exhibit behavior that emerges from large-scale training rather than from explicit design. Their competencies are distributed across high-dimensional parameter spaces rather than localized in symbolic rules; their outputs are probabilistic and context-sensitive rather than deterministic and context-independent.

This distinction is not merely technical. It signals an epistemological rupture: a qualitative change in what computation is and what it can meaningfully engage with. Conceptual frameworks inherited from symbolic AI are insufficient to account for systems whose behavior cannot be fully explained through rule inspection or algorithmic decomposition.

The tension between formal exactness and contextual judgment is not new - it predates generative AI by decades. In a celebrated exchange recounted by Arnold (2002), the physicist Zeldovich defined the derivative as the ratio of finite increments "assuming the latter is small" — deliberately omitting the apparatus of limits. The mathematician Pontryagin was outraged by this abandonment of logical rigor. Yet the same text reports the observation of the orbital ballistics specialist Lidov that the uniqueness theorem for differential equations — however impeccably proved — is a "mathematical fiction" in practice: the integral curves that mathematicians declare non-intersecting become physically indistinguishable at small scales. This is why, Lidov explained, ships cannot dock smoothly through automatic feedback control alone; the final approach requires either a sailor with a mooring line or rubber tires hung along the pier to absorb the inevitable impact. The anecdote illustrates a recurring pattern: formal systems capture a domain of explicit, verifiable truth, but practice inhabits a larger space where contextual judgment, approximation, and tacit know-how are indispensable. Generative AI, we argue, is the first computational paradigm that operates natively in this larger space.

It is important to distinguish this claim from a more familiar narrative. A common response to generative AI situates it within a progressive history of abstraction in programming: from machine code to assembly, from assembly to high-level languages, and from these to natural-language interaction. According to this view, generative AI merely represents a more accessible interface to an otherwise continuous paradigm. This continuity narrative is conceptually misleading. While previous increases in abstraction altered the expressiveness and convenience of programming languages, they preserved a constant epistemic structure: explicit specification followed by deterministic execution (Abelson & Sussman, 1996). Whether writing in assembly or Python, the programmer articulated procedures that the machine executed according to predefined rules. The output was, in principle, fully traceable to the specification. Generative AI breaks this invariant. The user does not specify procedures but steers emergent behavior; the output is not traceable to a specification but evaluated for contextual coherence. What changes is not the level of abstraction but the epistemic regime itself.

Recognizing this rupture — and resisting the temptation to domesticate it within the continuity narrative — is a necessary precondition for any coherent educational response.

## 2.2. From Explicit Knowledge to Operationalized Tacit Regularities

The distinction between explicit and tacit knowledge provides a critical lens for understanding the epistemic novelty of generative AI. Explicit knowledge can be fully articulated, formalized, and transmitted through symbolic representations. Tacit knowledge, by contrast, is embedded in practice, context, and judgment; it resists exhaustive formalization and cannot be fully decomposed into explicit rules (Polanyi, 1966; Ryle, 1949).

Traditional computing systems operate exclusively within the domain of explicit knowledge. Algorithms encode procedures that can, in principle, be completely specified; data structures represent information in fully formalized formats; program execution follows deterministically from stated premises. The expressive power of classical computation derives precisely from this explicitness.

Generative AI systems challenge this boundary. Through training on vast corpora of human-produced data, they acquire sensitivities to contextual, stylistic, and situational regularities that cannot be captured through explicit algorithmic specification alone. It is essential, however, to avoid anthropomorphic interpretations. Generative systems do not possess tacit knowledge in a phenomenological or experiential sense, nor do they "understand" in the human meaning of the term (Bender et al., 2021).

What they do provide is functional access to structures of practice that were previously accessible only through human tacit competence. Their epistemic novelty lies not in representing tacit knowledge per se, but in operationalizing sensitivity to context, nuance, and situated judgment. As a result, these systems can engage in forms of activity—interpretation, contextual adaptation, creative recombination — that fall outside the traditional scope of explicit computation.

This development does not invalidate explicit, rule-based computation, nor does it diminish its foundational role within computer science. Rather, it dissolves its exclusivity as the sole legitimate epistemic model of computational activity. Computation expands beyond the domain of fully articulable procedures into a plural landscape that includes operationalized forms of tacit regularity.

## 2.3. A Thought Experiment: Alien Contact and the Two Modes of Automation

The distinction between symbolic and vibe-based automation can be clarified through a thought experiment. Imagine attempting to communicate with an intelligent extraterrestrial civilization using two different technological aids. The first is a universal translator that maps symbols between languages based on pre-established correspondences — a system that presupposes shared semantic frameworks and fails when no common ontology exists. The second is a generative system that does not attempt translation but produces contextually coherent responses by analyzing the ongoing interaction, treating incoming signals not as expressions with fixed meanings but as perturbations to the current contextual state.

This contrast illustrates the core function of vibe-automation. Rather than applying predefined rules to symbolic inputs, a vibe-based system maintains and modulates a distributed representation of semantic and pragmatic context. Success is evaluated not in terms of correctness but coherence: whether the response sustains the interaction without disruption. In situations of radical uncertainty, participation precedes interpretation — shared meaning emerges as a product of sustained, coherent interaction, not its prerequisite (Wittgenstein, 1953). This resonates with Dreyfus's (1992) phenomenological critique of symbolic AI: expertise, in his account, is not the application of rules but the embodied capacity to respond appropriately to context — a capacity that generative systems now operationalize in their own, non-phenomenological way.

# 3. Technical and Cognitive Foundations of Vibe-Automation

Having established the epistemological foundations, the article now turns to the technical and cognitive dimensions of vibe-automation. The concept of vibe-automation, introduced above as a philosophical claim, has concrete technical substrates and demands specific cognitive competencies. This section examines three interconnected aspects: the transformation of training objectives from explicit metrics to tacit preference alignment (3.1); the emergence of a new professional figure — the Vibe-Engineer — whose cognitive posture differs fundamentally from that of the classical programmer (3.2); and the dissolution of algorithmic monopoly in favor of plural computational regimes that coexist within contemporary practice (3.3).

## 3.1. How the Loss Function Became a Vibe Check

The technical substrate of vibe-automation lies in the evolution of training objectives for AI systems. Classical machine learning optimized hard metrics: accuracy, precision, recall, F1 score---measures that are fully explicit and mathematically defined. The system either correctly classified an input or it did not; improvement was measured against objective ground truth (Bishop, 2006).

The introduction of Reinforcement Learning from Human Feedback (RLHF) represents a qualitative shift in this regime (Christiano et al., 2017; Ouyang et al., 2022). In RLHF, human evaluators do not provide explicit rules or correct answers. Instead, they express preferences--- selecting which of two outputs "feels better," is "more helpful," or "sounds more natural." These preferences encode precisely the kind of tacit, context-sensitive judgment that resists formalization: what reads as polite, what seems authoritative, what feels creative rather than generic.

RLHF is, in essence, the process of digitizing the human "vibe." The reward model trained on these preferences learns to approximate a function that maps outputs to an implicit space of human aesthetic and pragmatic judgment. The model optimized for this reward does not learn rules of good writing; it learns to navigate a landscape of felt qualities — a landscape with no explicit coordinate system but that is nonetheless navigable through gradient descent (Ziegler et al., 2019). While RLHF is paradigmatic, subsequent developments — Direct Preference Optimization (Rafailov et al., 2023), Constitutional AI (Bai et al., 2022), and Reinforcement Learning from AI Feedback — extend rather than supersede this logic: all encode the shift from explicit correctness criteria to distributed preference landscapes.

This is what distinguishes vibe-automation from classical automation at the technical level. The loss function is no longer a "check" of correctness against predetermined criteria. It is a "vibe check" — an evaluation of whether the output resonates with the distributed pattern of human tacit preferences. The optimization target has shifted from the space of verifiable propositions to the space of contextual coherence.

## 3.2. The Vibe-Engineer: From Conductor to Jazz Improviser

The advent of vibe-automation necessitates a fundamental reconsideration of the skills required for effective technological practice. For decades, engineering education has emphasized "algorithmic thinking" — the cognitive capacity to decompose complex problems into discrete steps, specify precise procedures, and reason about the behavior of deterministic systems (Wing, 2006). This mode of cognition, while valuable, is insufficient for the era of generative AI.

Consider the cognitive model implicit in classical programming. The programmer conceives an algorithm — a precise sequence of operations that, if followed exactly, will produce the desired output. They translate this mental model into code, a formal language that admits no ambiguity, no context dependence, and no room for interpretation. The programmer is like a composer writing a score: every note, every rest, every dynamic marking must be explicitly notated before a single sound is produced.

Working with generative AI requires a fundamentally different cognitive posture. The practitioner does not specify exact procedures but articulates intent, sets tone, and establishes the parameters of a generative process whose specific outputs cannot be predicted in advance. The better analogy is not the classical conductor but the jazz musician who sets a key, tempo, and feel, then responds in real-time to what the other players produce (Sawyer, 2007). When practitioners iterate on prompts to adjust tone or style, they are navigating latent space---searching for the region that corresponds to their tacit sense of what they want.

We propose the term Vibe-Engineer for this emerging professional figure. The Vibe-Engineer operates not through explicit specification but through resonance management: articulating intent at a high level, evaluating outputs through a combination of formal criteria and tacit judgment, and iteratively refining the collaboration between human purpose and machine capability. The core competency shifts from "doing things right" (efficiency within a known paradigm) to "doing the right-feeling things" (resonance within an emergent one).

## 3.3. From Algorithmic Monopoly to Plural Computational Regimes

Historically, the algorithm functioned as the central epistemic object of computer science, providing both technical solutions and a normative model of knowledge. While algorithms remain indispensable, generative AI dissolves their monopoly at the epistemological level.

Contemporary computational practice encompasses multiple regimes, each with distinct epistemic structures. The Symbolic regime operates through explicit rules and deterministic execution, validated by formal verification, with professional judgment centered on correctness of specification. The Statistical regime relies on probabilistic models and distributional assumptions, validated through statistical significance and confidence intervals, requiring judgment in model selection and assumption checking. The Neural regime involves learned representations and emergent features, validated via empirical benchmarks and ablation studies, demanding judgment in architecture choice and training strategy. The Generative/Vibe regime operates through operationalized tacit regularities and contextual coherence, validated by human evaluation, coherence assessment, and RLHF, requiring judgment in resonance management, taste, and contextual framing.

**Table 1. Plural Computational Regimes**

| Regime | Epistemic Structure | Validation Mode | Professional Judgment |
|---|---|---|---|
| **Symbolic** | Explicit rules, deterministic execution | Formal verification, proof | Correctness of specification |
| **Statistical** | Probabilistic models, distributional assumptions | Statistical significance, confidence intervals | Model selection, assumption checking |
| **Neural** | Learned representations, emergent features | Empirical benchmarks, ablation studies | Architecture choice, training strategy |
| **Generative / Vibe** | Operationalized tacit regularities, contextual coherence | Human evaluation, coherence assessment, RLHF | Resonance management, taste, contextual framing |

No single regime can be treated as universally normative. An adequate computer science curriculum must introduce students to this plurality and cultivate the ability to navigate across regimes---recognizing when algorithmic guarantees are appropriate, when probabilistic reasoning is required, when emergent behavior must be empirically validated, and when collaboration with opaque systems is justified by access to otherwise unreachable domains of practice.

## 4. The Two-Dimensional Proactive Education Framework

The epistemological analysis developed in the preceding sections demonstrates that the transformation facing computer science education is neither simple nor unidimensional. It operates simultaneously at multiple levels of depth — from foundational assumptions to institutional practices — and across multiple domains of action — from faculty identity to curriculum structure. To make this complexity tractable, the present section introduces a two-dimensional conceptual--operational matrix. The matrix is not a prescriptive blueprint but an interpretive architecture: it provides a structured means of identifying the specific tensions that must be addressed if educational transformation is to be coherent rather than fragmented.

The framework proposed here is explicitly architectural rather than linear. Its purpose is to make visible the multidimensional structure of the transformation facing computer science education. The framework is organized as a two-dimensional matrix that intersects epistemic depth with institutional action domains.

The vertical dimension represents three analytically distinct but interdependent levels of engagement: Level I — Conceptual Foundations (addressing the epistemological nature of the transformation); Level II — Transformation Principles (articulating what must change); and Level III — Institutional Implementation (specifying how changes can be enacted).

The horizontal dimension represents three pillars of institutional action: The Worldview Pillar (Faculty), focusing on epistemological orientation and professional identity; The Social Pillar (Industry and Profession), addressing the relationship between academia and evolving practice; and The Substantive Pillar (Curriculum), concerning content, structure, pedagogy, and graduate profiles.

The matrix should be read neither as a checklist nor as a linear sequence. It functions as a conceptual map: the same epistemic ideas reappear at different intersections because they operate differently at different levels and within different institutional domains.

**Table 2. The Two-Dimensional Transformation Matrix**

| | Worldview Pillar (Faculty) | Social Pillar (Industry & Profession) | Substantive Pillar (Curriculum) |
|---|---|---|---|
| **Level I: Conceptual Foundations** | Epistemological rupture; explicit vs. operationalized tacit regularities; limits of symbolic AI | Misinterpretation of GenAI as tool abstraction; erosion of stable professional identity | End of algorithmic monopoly; emergence of plural computational regimes; proactive education as paradigm |
| **Level II: Transformation Principles** | Faculty reconceptualization; differentiated epistemic roles; teaching as epistemic guidance | Shift from command to collaboration; bidirectional academia–industry partnership | Curriculum as navigational map; programming as epistemic infrastructure; Vibe-Engineer profile |
| **Level III: Institutional Implementation** | Faculty development programs; role redistribution; institutional support for epistemic change | Continuous dialogue with industry; joint sense-making of practice | Phased curriculum reform; outcome-oriented assessment; institutional learning loops |

## 5. Transformation Across the Three Pillars

The matrix introduced in the previous section provides a structural overview. The present section populates that structure with substantive analysis and concrete illustrative examples. Each of the three pillars — Worldview, Social, and Substantive — is examined across all three levels of the matrix (conceptual foundations, transformation principles, and institutional implementation). To ground the discussion in recognizable educational practice, each subsection concludes with an illustrative example that demonstrates how abstract principles translate into specific institutional decisions. These examples are not meant as universally applicable templates but as conceptual probes that clarify the framework's operational implications.

## 5.1. Worldview Pillar: Faculty Transformation and Epistemic Reorientation

### 5.1.1 Faculty at the Level of Conceptual Foundations

At the level of conceptual foundations, the primary challenge facing faculty is the need to recognize and internalize the epistemological rupture introduced by generative AI. Many educators in computer science have developed their professional identities within an algorithm-centric paradigm, where explicit specification, formal reasoning, and deterministic control constitute the normative model of legitimate knowledge. Within this paradigm, tacit or context-dependent forms of competence have traditionally been treated as peripheral or non-computational.

As a result, faculty may initially interpret GenAI as either a sophisticated automation tool or a threat to established standards of rigor. Both interpretations obscure the deeper transformation: the expansion of computation into domains characterized by operationalized tacit regularities. Faculty engagement at this level requires epistemic reorientation — an explicit confrontation with the limits of symbolic AI and an understanding of why plural computational regimes must now be treated as coexisting rather than hierarchical.

Example: Consider a senior algorithms professor who views GenAI primarily as a threat to academic integrity — a sophisticated cheating tool. This interpretation, while understandable, reflects an algorithm-centric epistemic position that equates computational legitimacy with explicit specification. Epistemic reorientation would involve recognizing that the professor's expertise in formal reasoning represents one regime within a plural computational landscape, and that their role evolves from sole authority to anchor of formal rigor within a broader ecosystem of computational modes.

### 5.1.2 Faculty at the Level of Transformation Principles

At the level of transformation principles, epistemic reorientation translates into changes in pedagogical stance. Teaching can no longer be conceived solely as the transmission of explicit procedures and formally verifiable solutions. Instead, it increasingly involves guiding students through environments characterized by partial opacity, probabilistic outcomes, and context-sensitive judgment.

This shift necessitates differentiation of faculty roles. Some educators continue to serve as anchors of formal rigor, maintaining the theoretical foundations — algorithms, complexity theory, formal methods — that remain indispensable. Others function as guides through epistemic transformation, helping students articulate when and why explicit guarantees give way to empirical validation or collaborative exploration with generative systems. Crucially, this differentiation should not be understood as a hierarchy but as an ecosystem of complementary roles (Boyer, 1990).

Example: In a redesigned software engineering course, one instructor handles formal requirements specification (symbolic regime), while another guides students through AI-assisted prototyping sessions where the task is not to write code but to evaluate, critique, and iteratively refine AI-generated solutions. Students learn that both modes of engagement represent legitimate computational expertise.

### 5.1.3 Faculty at the Level of Institutional Implementation

Effective transformation requires sustained opportunities for collective sense-making: seminars, reading groups, and structured dialogue focused on epistemological questions rather than mere tool adoption. Institutions must recognize that epistemic reorientation takes time and cannot be mandated through policy alone. Support mechanisms — such as adjusted workloads, recognition of diverse pedagogical contributions, and institutional patience — are essential.

Example: A department institutes a monthly "Computational Regimes Reading Group" in which faculty from different specializations jointly examine one artifact — a research paper, a generated codebase, or a failed AI deployment — through the lens of multiple regimes. An algorithms specialist and a machine learning researcher analyze the same system, making explicit where their epistemic assumptions diverge. Over time, these sessions build a shared vocabulary for discussing plural computation that feeds into collaborative course redesign.

## 5.2. Social Pillar: Industry Relations and the Transformation of Professional Identity

### 5.2.1 Professional Identity at the Level of Conceptual Foundations

At the level of conceptual foundations, generative AI challenges inherited assumptions about what it means to be a computer science professional. Historically, professional identity was anchored in the capacity to design, specify, and implement explicit algorithms whose behavior could be fully controlled and verified. The emergence of GenAI destabilizes this identity. The core epistemic shift — from control to collaboration, from procedure to judgment — reconfigures the boundaries of professional agency.

This transformation should not be interpreted as a devaluation of traditional expertise. Rather, it reflects an expansion of professional practice into domains where success depends on engaging with systems that have operational access to tacit regularities. Professional identity shifts from an exclusive reliance on algorithmic mastery toward a hybrid model that integrates formal rigor with contextual judgment.

Example: A software company hiring in 2025 increasingly seeks candidates who can demonstrate not only coding proficiency but also the ability to effectively collaborate with AI tools — specifying requirements in natural language, evaluating generated code for correctness and style, and orchestrating multi-agent workflows. The professional identity of a "developer" now spans both the symbolic and generative regimes.

### 5.2.2 Professional Identity at the Level of Transformation Principles

The dominant mode of interaction with computational systems shifts from command to collaboration. Professionals increasingly work alongside systems whose outputs must be interpreted, evaluated, and iteratively refined rather than simply executed (Brynjolfsson & McAfee, 2014). This shift has direct implications for how academia engages with industry. Traditional models of partnership — focused on tool alignment and short-term skill transfer — are insufficient. What is required is sustained, bidirectional sense-making: continuous dialogue aimed at understanding how computational practice evolves.

### 5.2.3 Industry Relations at the Level of Institutional Implementation

Concrete mechanisms must support continuous engagement between academic programs and professional practice: joint curriculum design initiatives, long-term partnerships centered on shared problem spaces, and institutionalized forums for reflection on evolving computational roles. Industry advisory boards should function not merely as evaluators of technical alignment but as partners in epistemic exploration---helping institutions anticipate shifts that cannot yet be fully codified into competencies or learning outcomes.

Example: A university establishes a "Computational Futures Lab" co-governed by faculty and industry partners. Rather than an annual advisory meeting, the lab conducts quarterly explorations of how GenAI is reshaping specific professional practices (e.g., code review, security auditing, system architecture). These explorations feed directly into curriculum revision cycles.

### 5.3. Substantive Pillar: Curriculum as Epistemic Navigation

### 5.3.1 From Ladder to Map: Curriculum at the Level of Conceptual Foundations

Traditional computer science curricula are structured as linear progressions---a ladder---from foundational mathematics through algorithms and data structures toward advanced applications. This structure presupposes delayed access to powerful tools and cumulative mastery of explicit knowledge.

Generative AI disrupts this assumption. Students now encounter advanced computational capabilities at the very beginning of their studies. The ladder model collapses: access to capability precedes understanding. A transformed curriculum must function as a navigational map rather than a ladder. Its purpose is to make visible the conceptual terrain of contemporary computation, including the tensions between explicit and operationalized tacit knowledge, determinism and emergence, verification and validation, control and collaboration.

Example: In the first semester, instead of beginning exclusively with data structures and algorithms, students encounter a "Computational Landscape" course that introduces all four regimes (symbolic, statistical, neural, generative) through hands-on exploration. They write a sorting algorithm, train a simple classifier, observe a neural network learning features, and interact with an LLM — learning from the outset that computation is plural, not monolithic.

### 5.3.2 Programming as Epistemic Infrastructure

Within this navigational model, programming remains essential, but its role is reconceptualized. Rather than serving as the defining craft of the discipline, programming functions as epistemic infrastructure — a medium through which computational systems are understood, interrogated, orchestrated, and evaluated (Papert, 1980).

Students must learn to program not primarily to produce code, but to develop conceptual literacy across computational regimes. This includes understanding when algorithmic specification is appropriate, when probabilistic modeling is required, when neural systems provide access to tacit regularities, and when agentic architectures support goal-directed interaction. Assignments should emphasize regime recognition, orchestration, and critical evaluation rather than isolated implementation.

Example: A capstone project requires students to build a system that integrates multiple computational regimes: a rule-based component for regulatory compliance, a statistical model for risk assessment, an LLM for generating human-readable reports, and an agentic layer for workflow orchestration. The assessment focuses not on the code per se but on the student's justification of regime choices, their evaluation methodology, and their reflective analysis of where each regime's limitations became apparent.

### 5.3.3 The Transformed Graduate Profile: The Vibe-Engineer

At the level of institutional implementation, the Substantive Pillar culminates in a redefinition of the graduate profile. The goal of the transformed curriculum is not the production of programmers defined by code-writing proficiency, but of Vibe-Engineers---computational professionals capable of sustained agency under conditions of epistemic uncertainty.

Such graduates demonstrate the ability to: recognize appropriate computational regimes for given problems; orchestrate hybrid systems spanning symbolic and generative paradigms; exercise judgment in the presence of opacity; evaluate outcomes through context-sensitive criteria that combine formal verification with tacit quality assessment; and maintain epistemic resilience as tools and paradigms continue to evolve.

Assessment practices must align with this profile. Evaluation shifts from the correctness of implementation toward the quality of judgment, the justification of methodological choices, and reflective awareness of limitations. In this way, the curriculum operationalizes the principles of proactive education within everyday pedagogical practice.

## 6. The Beige Box: Mode Collapse and Cultural Homogenization

The preceding sections have addressed the epistemological, technical, and institutional dimensions of the transformation that generative AI introduces into computer science education. But vibe-automation carries a risk that is neither purely technical nor purely institutional — a risk that concerns the cultural and aesthetic quality of what generative systems produce and, by extension, of what students learn to accept as adequate. If the previous sections asked how to educate about vibe-automation, this section asks what happens when vibe-automation goes wrong. It argues that the cultivation of taste — specific, culturally grounded aesthetic judgment—constitutes not merely a desirable educational outcome but a necessary form of resistance to what we call generative homogenization.

## 6.1. The Averaging Problem

The emergence of vibe-automation introduces a risk that is neither purely technical nor purely cultural but irreducibly both. Generative AI systems trained through RLHF optimization tend to converge on outputs that satisfy aggregate human preferences — producing text that is competent but undistinctive, images that are attractive but generic, code that is functional but architecturally uninspired. This phenomenon parallels mode collapse in generative adversarial networks, in which the system collapses into producing a narrow subset of possible outputs (Goodfellow et al., 2014).

We call this the "Beige Box" problem: the tendency of optimized generative systems to converge on a bland aesthetic center — the color that emerges when all colors are averaged. The risk is particularly acute given the feedback dynamics of AI-generated content. As more of the internet's content is produced by generative AI and that content becomes training data for future models, we risk a recursive loop of homogenization in which each generation drifts further from the full diversity of human expression (Shumailov et al., 2024).

## 6.2. Taste as Resistance

This is why the cultivation of taste becomes not merely a practical skill but an ethical imperative. The practitioner with a developed taste can recognize and resist the pull toward beige — can detect when outputs are technically adequate but aesthetically impoverished, and can demand and achieve distinctiveness in a world of generated uniformity. The curator with deep knowledge of jazz can use generative tools to extend that tradition rather than dilute it. The designer rooted in a particular visual culture can guide a generation toward outputs that honor rather than homogenize that heritage.

Education for Vibe-Automation must therefore cultivate not just generic aesthetic competency but specific cultural fluencies — deep familiarity with particular traditions, styles, and forms of excellence. This is why the humanities within technical universities — rhetoric, logic, ethics, art history — are no longer "electives" but the foundation for crafting effective prompts and evaluating results with genuine discrimination. The necessity of emphasizing the humanities transforms from a cultural nicety to an economic and epistemic imperative (Nussbaum, 2010).

Example: A course titled "Computational Aesthetics" requires students to develop expertise in a specific cultural tradition (e.g., Japanese woodblock printing, West African textile patterns, Bauhaus typography) and then use generative tools to produce work that extends rather than dilutes that tradition. Assessment is conducted by domain experts who evaluate whether the AI-assisted output demonstrates genuine understanding of the tradition's principles, or merely surface-level mimicry.

## 7. Conclusion

The transformation of computer science education in the era of generative artificial intelligence cannot be addressed through incremental curricular updates or isolated technological interventions. As this framework has argued, the emergence of GenAI constitutes an epistemological rupture that alters the nature of computation itself, expanding it beyond the exclusive domain of explicit, rule-based knowledge and into regimes characterized by operationalized tacit regularities, contextual judgment, and probabilistic emergence.

The concept of Vibe-Automation of Automation captures this shift with precision that the original formulation — "automation of automation" — can no longer provide. Where Domingos described machines learning to program, we now witness machines learning to resonate, to navigate contextual spaces that resist explicit articulation. The Vibe-Engineer emerges as the professional figure suited to this new landscape: not a displaced programmer but a computational professional whose expertise lies in orchestrating multiple regimes, exercising judgment amid opacity, and maintaining the cultural and epistemic specificity that generative averaging threatens to dissolve.

The two-dimensional framework developed here — intersecting levels of epistemic depth with pillars of institutional action — provides a means of holding together conceptual understanding, transformation principles, and concrete implementation without collapsing one into another. The Worldview Pillar insists that no curricular transformation is possible without faculty epistemic reorientation. The Social Pillar demonstrates that professional identity and industry relations must be reconceptualized as sites of shared sense-making. The Substantive Pillar translates these insights into curricular architecture, redefining programming as epistemic infrastructure and the graduate profile in terms of judgment, orchestration, and resilience.

Together, these pillars articulate a vision of proactive education oriented toward sustained agency under conditions of structural uncertainty. Computer science education neither retreats into nostalgic defense of algorithmic purity nor dissolves into uncritical adoption of generative tools. Instead, it reasserts its intellectual integrity by embracing epistemic pluralism and by equipping students to navigate the complexity of contemporary computational practice.

The framework presented here is not a finished blueprint but an invitation to institutional experimentation grounded in conceptual clarity. Several limitations should be acknowledged. The framework is conceptual rather than empirical: it has not been tested through implementation in an actual curriculum reform process, and the illustrative examples, while grounded in recognizable educational practice, are not drawn from longitudinal case studies. The analysis is situated primarily within the context of Western, English-language computer science education; its applicability to other educational traditions and institutional cultures requires separate investigation. The treatment of industry relations remains schematic; a fuller account would need to engage with the political economy of technology education and the power asymmetries between academic institutions and technology corporations. Finally, the framework addresses undergraduate education; its implications for graduate research training and continuing professional development remain to be explored.

Its relevance extends beyond computer science, offering a paradigmatic case for how academic disciplines may respond to technological transformations that challenge their foundational assumptions. In this sense, proactive education emerges not as a reaction to generative AI but as a necessary mode of academic self-renewal in an era where the boundaries of computation, knowledge, and professional agency are being fundamentally redrawn.